\documentclass[11pt]{article}

\usepackage[final]{acl}

\usepackage{times}
\usepackage{latexsym}

\usepackage[T1]{fontenc}

\usepackage[utf8]{inputenc}

\usepackage{microtype}

\usepackage{inconsolata}

\usepackage{graphicx}

\usepackage{amsmath}
\usepackage{amsfonts}
\usepackage{subfiles}
\usepackage{tcolorbox}

\title{Narrowing the Horizon: Quantifying Topic Saliency Shifts in Generative Monoculture}

\author{Oriane Peter \and Elena Simperl \and Kate Devlin \\
  King's College London \\
  London, United Kingdom \\
  \texttt{oriane.peter@kcl.ac.uk}}

\begin{document}
\maketitle
\begin{abstract}
As Large Language Models (LLMs) become central to how we access and share information, they play an increasingly powerful role in shaping global knowledge.  However, as these models evolve, their outputs risk converging into a \textit{generative monoculture}, where the diversity of perspectives they represent narrows over time. Studies at the model level often fail to pinpoint which specific topics or viewpoints are being marginalised or amplified in this process. In this paper, we introduce a method to measure shifts in topic saliency across model families, tracking what gains or loses prominence during post-training. Applying this approach to a case study of climate change discourse, we demonstrate how homogenisation affects the representation of diverse solutions across different models. We also test interventions to counter this trend, showing that specialised models can help preserve a broader range of perspectives. This underscores the importance of monitoring topic saliency to diagnose the risks of monoculture and to ensure AI systems reflect a pluralism of ideas. Data and Code are accessible \href{https://github.com/oriane/topic_saliency_shift}{here}.

\end{abstract}

\section{Introduction}
Large Language Models (LLMs) mediate an increasing share of global knowledge production, from academic writings \cite{liangMappingIncreasingUse2024}, journalistic outputs \cite{thurmanAIAdoptionUK2025} to Wikipedia articles \cite{brooksRiseAIGeneratedContent2024}. Concomitantly, recent empirical research suggests that outputs across disparate model families are converging toward a narrow, homogenous subset of representations, a phenomenon coined \textit{generative monoculture} \cite{wuGenerativeMonocultureLarge2024}. The intersection of this monoculture with the massive scale of LLM deployment raises concerns regarding the collapse of epistemic diversity, i.e., the plurality of knowledge accessible  \cite{petersonAIProblemKnowledge2025}. Such a global loss of diversity may have far-reaching consequences, potentially eroding our collective capacity to address multifaceted challenges where a pluralistic stance is key. 


While prior work has investigated homogenisation at various scales, from internal model representations \cite{guoBenchmarkingLinguisticDiversity2024, hamiltonDetectingModeCollapse2024, murthyOneFishTwo2025, sorensenPositionRoadmapPluralistic2024} to downstream impacts on users \cite{chenSpiralSilenceHow2024, padmakumarDoesWritingLanguage2024, wangLargeLanguageModels2025, souratiHomogenizingEffectLarge2026} and society \cite{wuGenerativeMonocultureLarge2024, gillespieGenerativeAIPolitics2024, bommasaniPickingSamePerson2022}, studies at the model level often fail to robustly identify the specific topics or perspectives around which models converge. This lack of resolution prevents a concrete assessment of the consequences of generative monoculture. Rather than treating homogenisation as an undifferentiated loss of diversity, identifying which information is lost is therefore necessary for a nuanced analysis of the resulting societal impacts. We argue that this resolution can be obtained by quantifying shifts in topic saliency across model families. The goal of this work is therefore not to propose an \textit{alternative} to existing homogenisation measurements, but rather a \textit{complementary} analysis at the topic level. Since these impacts are most evident when grounded in specific contexts, we propose a methodology for quantifying shifts in topic saliency and apply it to the domain of climate change discourse. This approach moves beyond theoretical diversity loss to identify concrete societal risks and actionable, model-level mitigation strategies. We thus make three contributions: 
\begin{itemize}
\item A methodology for quantifying shifts in topic saliency during LLM homogenisation;
\item  Empirical evidence demonstrating that tracking topic saliency is essential for evaluating the impact of generative monoculture and the efficacy of specific interventions, within the context of climate change discourse; and
\item An evaluation of mitigation strategies, providing empirical support for the use of specialised models in mitigating the effects of output convergence.
\end{itemize}
\section{Background}

\subsection{LLM Homogenisation} \label{sec:homo}
LLMs have been empirically shown to tend towards generative monoculture, or the ``significant narrowing of model output diversity relative to available training data for a given task'' \cite{wuGenerativeMonocultureLarge2024}. This effect leads to highly repetitive outputs with lower semantic diversity than would be generated by humans writing \cite{jiangArtificialHivemindOpenEnded2025, wuGenerativeMonocultureLarge2024, guoBenchmarkingLinguisticDiversity2024, gillespieGenerativeAIPolitics2024, murthyOneFishTwo2025} or simple web searches \cite{wrightEpistemicDiversityKnowledge2026}. This homogenisation extends beyond individual model families, with model outputs increasingly resembling one another \cite{jiangArtificialHivemindOpenEnded2025, wuGenerativeMonocultureLarge2024, kimCorrelatedErrorsLarge2025}. Furthermore, this diversity loss worsens when models are trained on synthetic output \cite{hamiltonDetectingModeCollapse2024, schaefferPositionModelCollapse2025, shumailovAIModelsCollapse2024, guoCuriousDeclineLinguistic2024, alemohammadSelfConsumingGenerativeModels2023} and has a homogenising effect on users \cite{chenSpiralSilenceHow2024, padmakumarDoesWritingLanguage2024, parkDiminishedDiversityofthoughtStandard2024, souratiHomogenizingEffectLarge2026, srivastavaLargeLanguageModels2025}, suggesting the emergence of a feedback loop which may increase this narrowing over time.

Whilst there are many factors contributing to models reproducing only a subset of their training diversity, ranging from training batch composition \cite{farnadiPositionCrackingCode2024} to decoding algorithm choice \cite{holtzmanCuriousCaseNeural2019, pavlovicUnderstandingEffectTemperature2024, dhamalaAnalysisEffectsDecoding2023}, post-training has been most consistently identified as the main driver of output flattening \cite{casperOpenProblemsFundamental2023, kirkUnderstandingEffectsRLHF2024, sorensenPositionRoadmapPluralistic2024, omahonyAttributingModeCollapse2024, hamiltonDetectingModeCollapse2024, fengModularPluralismPluralistic2024, murthyOneFishTwo2025}. 

\subsection{Epistemic Collapse} \label{sec:epistemic}

Output homogenisation is not an issue in and of itself. For example, after post-training, models should generate texts which are less toxic, more accurate, and better aligned with the user's request, all of which may reduce the set of likely outputs. However, this homogenisation does become undesirable when it threatens humans' epistemic diversity. \citet{petersonAIProblemKnowledge2025} describes this phenomenon as \textit{Knowledge Collapse}: a ``progressive narrowing over time (or across technological iterations) of the set of information available to humans''. This narrowing is brought both by generative monoculture and the increasing roles of LLMs systems as `Knowledge Distribution Systems': infrastructures that mediate epistemic resources at scale \cite{lavazzaGenerativeAIKnowledge2026}. AI-mediated knowledge distribution, from web search summarisation to chatbots, is becoming comparatively easier to access than alternatives, such as libraries. Consequently, humans are increasingly unlikely to encounter perspectives outside the narrow set promoted by these systems, or the even narrower echo-chambers which the personalisation of AI systems can lead to \cite{kirkPersonalisationBoundsRisk2023a, sharmaGenerativeEchoChamber2024}. The combination of the scale of their deployment and generative monoculture leads, in effect, to a shrinkage of the set of points of view replicated, and, potentially, to a collapse of accessible knowledge. 

Furthermore, the knowledge that remains accessible serves an `agenda-setting' function: the salience of accessible knowledge influences public attitudes and thus shapes political agendas \cite{mccombsAgendaSettingFunctionMass1972, bhatReactiveWritersHow2026}. While mass media exerted similar effects before the advent of digital resources, these new technologies have shifted the centralised authority of media toward a complex network of influences, shaped by algorithmic recommender systems, user-generated content, and search engines \cite{almakatyAgendaSettingTheory2025}. When only a few LLMs function as a centralised `Knowledge Distribution System', the authority to determine what knowledge is propagated becomes once again concentrated in a few dominant institutions, shifting informative and interpretive power toward often private actors and interests \cite{gur-ariehAmbiguityCollapseLLMs2026, lavazzaGenerativeAIKnowledge2026, peterDecentralisingLLMAlignment2025}.

\subsection{Homogenisation and Pluralism}

Therefore, to understand the implications of generative monoculture, it is necessary to characterise exactly what is lost and \textit{which} knowledge remains after collapse, as these remnants outline the new boundaries of public discourse and the resulting shifts in political power. 

However, current research into homogenisation is primarily descriptive; it succeeds in quantifying the extent of the effect but stops short of qualifying its nature. For example, \citet{jiangArtificialHivemindOpenEnded2025} identifies both \textit{intra-model repetition} and \textit{inter-model homogeneity} across a wide array of real-world queries. Similarly, \citet{wrightEpistemicDiversityKnowledge2026} empirically demonstrates that across 27 models, LLM outputs produce a significantly narrower set of claims compared to even basic web searches. While both works effectively illustrate a systemic loss of diversity, they provide little insight into which specific perspectives or cultural frameworks endure.

On the other hand, research on LLMs \textit{pluralism} does focus on whose opinions are being left out. These approaches are concerned with identifying \textit{which humans} or \textit{whose values} are present in a model's outputs, clustered across sociodemographic lines, such as nationality \cite{durmusMeasuringRepresentationSubjective2023, sorensenPositionRoadmapPluralistic2024, santurkarWhoseOpinionsLanguage2023, atariWhichHumans2023} or political orientation \cite{wrightLLMTropesRevealing2024, fengPretrainingDataLanguage2023, rozadoPoliticalPreferencesLLMs2024, rottgerIssueBenchMillionsRealistic2026}. They have demonstrated that LLMs are most often reflective of the perspective of a small, privileged subgroup \cite{atariWhichHumans2023, durmusMeasuringRepresentationSubjective2023}, an effect which is also intensified by post-training \cite{sorensenPositionRoadmapPluralistic2024}. However, these methods seek to identify the \textit{bias} of the model, or the leaning expressed stance, rather than the breadth of topics or concepts elicited in the first place. 

Consequently, neither homogenisation nor pluralism offers an analytical investigation of topic-level shifts in LLMs' homogenisation. We propose to do so by complementing diversity studies with a quantification of pluralism, measured as shifts in topic saliency. Accordingly, we develop a robust, free-text-based methodology that identifies which types of perspectives are prioritised and which are 'smoothed away' during post-training. For a more detailed discussion of how our methodology relates to and diverges from existing homogenisation and pluralism benchmarks, see \autoref{sec:detail_dis}.

\subsection{The Wicked Problem of Climate Change}

To ground our analysis in concrete societal implications of homogenisation, we focus on the `wicked' problem of climate change. `Wicked problems' are complex but existential risks that lack singular definitions or solutions. These challenges resist reductionist approaches, requiring instead interdisciplinary frameworks to explore the many facets of the issue \cite{lonngrenWickedProblemsMapping2021, incroperaClimateChangeWicked2015}. Therefore, epistemic collapse severely undermines our collective ability to address such challenges: By flattening the public discourse and accessible information, these models may fail to support the plurality of perspectives necessary to address global environmental crises and exert a dangerous narrowing `agenda-setting' influence on policy priorities \cite{mccombsAgendaSettingFunctionMass1972, trielli2022algorithmic}. Consequently, we analyse the homogenisation of climate solutions to understand how LLMs shape the collective understanding of viable interventions.

\section{Methodology: Quantifying Topic Saliency Shifts}

Our proposed methodology to identify social harms by measuring shifts in topic saliency satisfies several desiderata. Given that demonstrating \textit{systematic} knowledge suppression requires identifying a convergent narrowing \textit{trend}, it offers an analysis across diverse model families. Additionally, it allows for distinguishing between preserved and suppressed topics in order to draw explicit conclusions on the consequences of topic set shrinking. Finally, it provides insight into how post-training interventions trigger these shifts, thereby informing mitigation strategies.

We use \textit{base} models (pre-trained only) as our baseline for measuring homogenisation, meaning the signal available before the diversity reduction resulting from post-training (see \autoref{sec:homo}) \footnote{ We consider post-training generally in this paper as the \textit{ft} version released by institutions on Hugging Face, which typically encompasses instruction tuning, supervised fine-tuning, and alignment procedures.}. While a real-world corpora might introduce uncontrollable confounds such as sample selection or audience intent, this baseline isolates the homogenisation effects under controlled, reproducible conditions on identical prompts. Furthermore, because the \textit{base} models are already a compression of pre-training data, which are themself a filtered version of human knowledge,  we argue that this reference point is conservative: any measured homogenisation likely underestimates the true narrowing effect relative to human diversity of opinion. Therefore, we define homogenisation as the contraction of the topic set represented in a model's output distribution relative to its \textit{base} baseline. 

Using a pre-trained model as a baseline requires safeguarding against its tendency toward task-irrelevant noise, which would falsely inflate baseline diversity. We mitigate this by applying a targeted key term extraction pipeline to isolate domain-relevant concepts, grounding our analysis in the specific context of focus.

This baseline allows for a cheap and controlled comparison of the variation in topic prominence in the model output, caused by diverse design interventions. This baseline does not, however, provide insight into the validity or quality of the surfaced topics, which must be compared with the relevant literature, an analysis we provide in the discussion.  

In this paper, we test the effect of post-training, increased parameter count, using leading (if not state-of-the-art) models, training specialised models, and prompt engineering on the frequency of mention of different climate solutions. 

\subsection{Models}
Our evaluation includes three variants across eight open-source model families: \textit{Apertus, Gemma 2, Gemma 3, Llama 3, Mistral, OLMo 3, Qwen 2.5,} and \textit{Qwen 3}. For each family, we evaluate the \textit{base} and post-trained version of the model closest to ~8B parameters. We also include a variant with the highest available parameter counts (ranging from 27b to 80b) to control for scaling effects. Additionally, we benchmark against efficient and cost-optimised versions of leading closed-source models, specifically \textit{Claude Haiku, Llama 4, GPT-5-Nano, and Grok 4.1 Fast}, which represent the latest fast-inference variants available from major developers as of early 2026. We refer to this model as 'leading' in the rest of the paper \footnote{We refrain from using actual state-of-the-art models due to the prohibitive cost of repeated outputs sampling.}.  Finally, we also test a model specifically fine-tuned to answer climate-related questions \textit{ClimateGPT} \cite{thulkeClimateGPTAISynthesizing2024}, see \autoref{sec:appendix-models} for more details.

\subsection{Prompting}
 We elicit a broad set of concepts via open-text prompting through a set of 64 paraphrased prompts with similar meaning but varying linguistic structure, such as \textit{``In a sentence, describe the \{best / most promising\} concrete \{approach / solution\} to \{climate adaptation / resolve the climate crisis\} .''} (see \autoref{sec:appendix-prompts} for all combinations). We deliberately use ambiguous phrasing to test how models converge on a narrow set of interpretations of terms like `best' or `efficient'. Following the sampling protocols of \citet{jiangArtificialHivemindOpenEnded2025} and \citet{wuGenerativeMonocultureLarge2024}, we sample each prompt n=50 times with a temperature T=1.0 and top-p=0.9.  This procedure yielded a total of 92'800 distinct completions (64 prompts x 50 iterations x 29 models). 
 
 Furthermore, we test a Chain-of-Thought (CoT) version of the prompt on leading models, following observations by \citet{meinckePromptingDiverseIdeas2024} that CoT prompting can widen the diversity of generated ideas. We adapted the product-design-based CoT prompt from \citet{meinckePromptingDiverseIdeas2024} to fit our use case, first asking the model to `brainstorm' maximally diverse solutions before answering the same core question used in our base prompting strategy, see \autoref{sec:cot} for the details. To ensure a direct comparison with the base results, evaluation is performed exclusively on this final output. This generates (64 prompts x 50 iterations x 4 models) 12'800 extra completions, leading to a total of 105'600 data points. 

\subsection{Topic Extraction}
To identify which solutions are covered by the model responses, we employ a two-stage exploratory topic extraction pipeline. First, we extract representative keywords from the free-text completions using an LLM-as-a-judge framework \citep{mansourHowWellLarge2025}. To ensure the quality of the keyword extraction step, we conducted a manual validation experiment. We randomly sampled and manually annotated 195 outputs (5 prompts for each model setup). We then compared \texttt{gpt-5-mini} against this ground truth and the best-performing model from \citet{mansourHowWellLarge2025}: \texttt{LLaMA 3.1 70B}. The \texttt{gpt-5-mini} judge achieved a 0.84 F1 fuzzy matching score and a 0.94 recall, strongly outperforming Llama (0.75 F1, 0.7 recall), ensuring that our pipeline rarely produces false negatives. We also verified that the missed keywords did not systematically belong to the same topic; see \autoref{sec:topic-extraction} for more details. For all subsequent experiments, we therefore extracted keywords using \texttt{gpt-5-mini}, yielding 95'568 unique keywords across the climate corpus. 

The keywords were then grouped into topics through a hybrid annotation strategy, which provides a more robust ground truth than fully automated classification \citep{baumannLargeLanguageModel2025} by balancing two competing objectives: maintaining an exploratory framework to avoid imposing researcher-defined preconceptions of climate solutions, while simultaneously mitigating the risk of propagating the inherent biases of the models under study.

Our pipeline operates in three stages. First, we randomly select a representative subset of 12,000 keywords across all model outputs and prompts. We perform an exploratory pre-grouping of these keywords by clustering their semantic embeddings using HDBSCAN \citep{campelloDensityBasedClusteringBased2013}. Second, these initial clusters are manually curated and assigned topical labels; this ensures the resulting taxonomy is grounded in human-verified categories rather than purely algorithmic groupings. Finally, this curated subset serves as the gold-standard reference for categorising the remaining 83'568 keywords via a k-Nearest Neighbour (k-NN) classification within the same embedding space. All embeddings are obtained through the OpenAI \textit{text-embedding-3-small} model, see \autoref{sec:topic-classification} for the details.

This workflow produced 48 distinct climate-related topics. To validate the reliability of this pipeline, we performed manual verification on a balanced test set, achieving an accuracy of 89\%. Representative keyword lists for each topic and examples of associated answers are provided in \autoref{app:climate_keywords}.

\subsection{Metrics}
\paragraph{Quantifying Homogeneity} To ensure we measure genuine semantic diversity rather than variations stemming from irrelevant or low-quality "noise", which is more frequent in pre-trained models, we compute the diversity of the keywords rather than comparing full outputs, as keywords are already filtered for relevance. We rely on the mean cosine distance between the keywords' embeddings, which is a commonly used measure of diversity \cite{jiangArtificialHivemindOpenEnded2025, omahonyAttributingModeCollapse2024, guoBenchmarkingLinguisticDiversity2024, guoCuriousDeclineLinguistic2024},  using OpenAI's \textit{text-embedding-3-small} embeddings.

Formally, for each model $m$ and prompt $p$, we pool all keywords extracted across the $N=50$ outputs into a single list $L_{m,p} = \{w_1, \dots, w_K\}$ where $K$ is the total number of keywords elicited by prompting $m$ with $p$. The \textit{semantic spread} $\sigma_{m,p}$ is defined as the mean pairwise cosine distance across this aggregate list:

\begin{equation}
\sigma_{m,p} = \frac{1}{\binom{K}{2}} \sum_{i < j} d_{cos}(w_i, w_j)
\end{equation}

Note that we use the full list of keywords rather than the set of distinct keywords, as repeated keywords across trials show lower diversity, thus capturing both conceptual narrowing and lexical repetition.

To evaluate the performance of different model types $\mathcal{T}$ (such as \textit{base} vs \textit{post-trained} models), we define the Empirical Cumulative Distribution Function (ECDF) over the set of all observed spreads:

\begin{equation}
F_{\mathcal{T}}(x) = \frac{1}{|M_{\mathcal{T}}| |P|} \sum_{m \in M_{\mathcal{T}}} \sum_{p \in P} \mathbb{I}(\sigma_{m,p} \le x)
\end{equation}

where $M_{\mathcal{T}}$ represents the set of models belonging to type $\mathcal{T}$, $P$ is the set of prompts, and $\mathbb{I}$ is the indicator function. A rightward shift in $F_{\mathcal{T}}(x)$ signifies that a model type consistently produces a higher degree of semantic diversity.

\paragraph{Quantifying Topic Saliency Shifts}

To estimate and isolate the effect of post-training on topic prevalence, we define a Bayesian Binomial Generalised Linear Mixed Model (GLMM). For each topic $k$, the number of occurrences $y_{ki}$ out of $N=50$ samples for a given model and prompt is modelled as:

\begin{equation}
    y_{ki} \sim \text{Binomial}(N, p_{ki})
\end{equation}
\begin{equation}
     \text{logit}(p_{ki}) = 
    \alpha + \sum_{j \in \{FT, Large\} }{\beta_j \cdot X_j} +  \gamma_{m[i]}  + \delta_{q[i]}
\end{equation}

where $\alpha$ is the intercept representing the baseline log-odds in the \textit{base} state; $\beta_j$ are the fixed effects for post-trained and large (higher parameters count) variants; $\gamma_{m[i]} \sim N(0, \sigma^2_m)$ is the random effect for the model family; and $\delta_{q[i]} \sim N(0, \sigma^2_q)$ is the random effect for the specific prompt variation.

This framework accounts for random effects from both individual model families and specific prompt formulations, thereby controlling for inherent model-specific biases and the varying propensity of certain prompts to elicit specific concepts. By treating the transition from \textit{base} to \textit{post-trained} (with both similar and increased parameter counts) as a fixed effect, we statistically isolate the directional impact of post-training alignment on topic prevalence. We utilised the \textit{Bambi} library \cite{caprettoBambiSimpleInterface2022} for Bayesian estimation, identifying credible shifts in the latent distribution across 48 topics.

\paragraph{Comparing Mitigation Approaches}
To assess how mitigation strategies recover topic-level diversity, we compare the output density profiles of a specialised model (ClimateGPT) and four popular leading models (prompted with both vanilla and CoT approaches). While GLMMs offer robust evaluation of post-training effects, they require access to base model states, which is not possible for proprietary leading models. Given their global ubiquity \cite{aubakirovaStateAIEmpirical}, understanding the topic prevalence of these models is also critical to identifying generative monocultures. We therefore utilise Kernel Density Estimation (KDE) to analyse the distribution of prompt-level win rates. Each win rate represents the proportion of $N=50$ repetitions in which a topic was mentioned for a specific model-prompt pair. By contrasting these distributional shapes, we can compare the topic saliency shift between model types, with a left shift indicating suppression and a right one promotion.

\subsection{Assessing Generalisation}

Our core analysis focuses on climate change to ensure methodological rigour and the domain grounding and depth necessary to identify the impacts of homogenisation. However, to evaluate generalisability, we apply our framework to a second wicked problem: homelessness and poverty \cite{rittelDilemmasGeneralTheory1973, gruendelTechnopoliticsWickedProblems2022}.

Specifically, we evaluate the same 8 \textit{base} models alongside their corresponding \textit{post-trained large} variants using 64 domain-adapted prompt variations (detailed in \autoref{sec:appendix-prompts}). Across 51,200 generated completions (64 prompts × 50 iterations × 16 models), we extracted 23,096 keywords. From these, 1,314 keywords were clustered and manually curated into a reference set for K-NN classification, resulting in 64 distinct categories. Representative keyword lists are available in \autoref{tab:poverty_keywords}.

\section{Results}
\subsection{Quantifying Homogeneity} \label{sec:homo_res}
\begin{figure}[hbt!]
    \centering
    \includegraphics[width=\columnwidth]{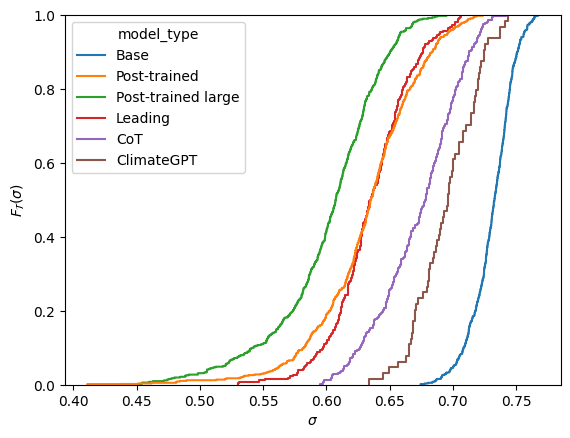}
    \caption{\textbf{Empirical Cumulative Distribution Function (ECDF) ($F_{\mathcal{T}}(\sigma)$) of intra-prompt semantic spread ($\sigma$)}. Higher values (shifts to the right) of $\sigma$ indicate greater semantic diversity within the model’s output distribution.}
    \label{fig:sementicspread}
\end{figure}

As expected, \autoref{fig:sementicspread} shows that the \textit{base} versions consistently retain the highest diversity across all models and variations, validating their use as a baseline. Additionally, a comparison between the post-trained and large post-trained versions of the same models reveals that while post-training diminishes output diversity, increasing model scale appears to exacerbate this homogenisation. Furthermore, while leading models mitigate some of the homogenisation observed in larger previous-generation models, they do not offer higher diversity than smaller post-trained variants. Notably, both CoT prompting and model specialisation increase keyword heterogeneity; the specialised \textit{ClimateGPT} model, in particular, appears to conserve the highest degree of semantic diversity.

\subsection{Quantifying Topic Saliency Shifts} \label{sec:characterisation}

\begin{figure*}[hbt!]
    \centering
    \includegraphics[width=0.8\textwidth]{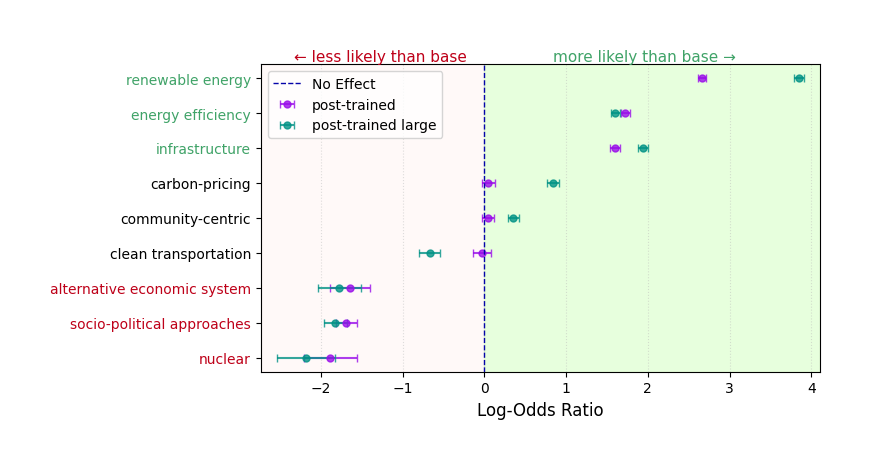}
    \caption{\textbf{Bayesian GLMM Credible Intervals for topic frequency shifts.} The plot displays the top, middle, and bottom three topics, ordered by the log-odds ratio of the post-training effect. Intervals are shown for the post-training (purple) and the post-training large (green) model effect relative to a pre-trained baseline. Topics in green indicate a credible \textit{increase} in prevalence, black \textit{no effect}, and red \textit{suppression}. The log-odds ratio represents the shift in likelihood compared to the baseline (N.B. a log-odds of 4 indicates an $\exp(4) \approx 54$ times more likely appearance.)}
    \label{fig:glmm}
\end{figure*}

\autoref{fig:glmm} illustrates the GLMM results for the top, middle, and bottom three topics, ranked by the log-odds ratio of the post-training effect. Our results demonstrate that certain topics, such as \textit{Renewable Energy}, are significantly amplified by post-training. This effect is further exacerbated by increasing model size; for example, a log-odds ratio of 3.8 indicates that this topic is approximately $e^{3.8} \approx 44.7$ times more likely to appear in the output of a large model than in the smaller \textit{base} version.

Conversely, topics related to systemic change, such as \textit{Socio-political Approaches} or \textit{Alternative Economic Systems}, are credibly suppressed following post-training. Notably, the mention of \textit{Nuclear Energy} also becomes significantly less likely post-alignment. The complete list of effects for all 48 topics is provided in \autoref{fig:glmm_full}.

Qualitatively, we can see that the base models provide answers coherent with the prompt, and that post-trained outputs tend to be more concise and technical answers. For example, when asked about the \textit{``best concrete pathway to resolve the climate crisis''}, a pre-trained version of the google \textit{gemma-2-9b} model's outputs include plausible-sounding answers such as: \textit{``The pathway we need is a social and environmental transformation at a societal level, starting with the realization that climate change is also a social and economic justice problem, then going on to develop and implement strategies for tackling the many dimensions of the climate crisis, from energy supply to mobility and agriculture, while prioritizing a fair and just transition for all stakeholders in the process.''}, while the post-trained version gives answers such as: \textit{``Rapidly transitioning to renewable energy sources while simultaneously implementing ambitious carbon capture and storage technologies is the most effective way to mitigate the climate crisis.''}. More qualitative examples can be found in \autoref{tab:output_examples}.

\subsection{Comparing Mitigation Approaches} 

\begin{figure*}[hbt!]
    \centering
    \includegraphics[width=\textwidth]{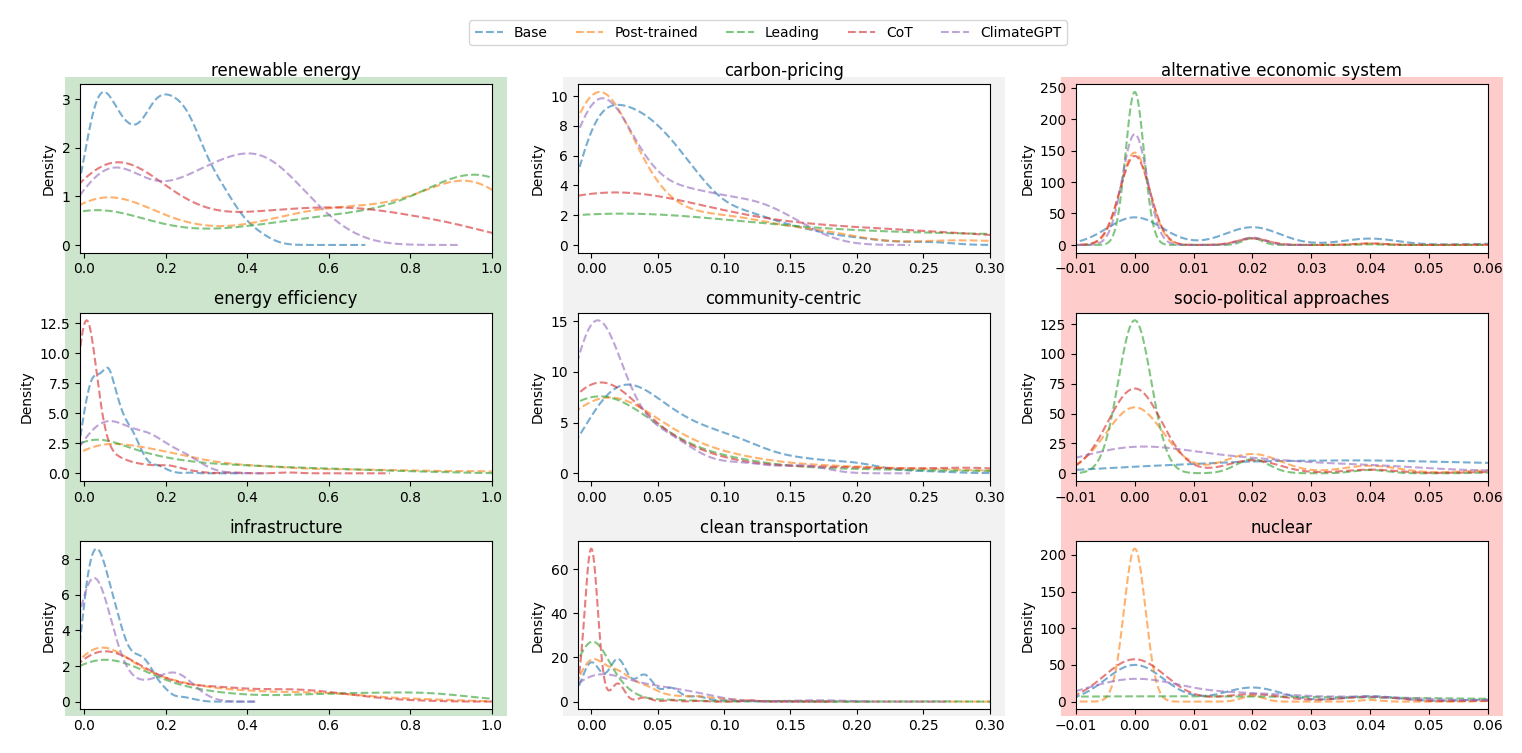}
    \caption{\textbf{Distribution of Topic Saliency}. The Kernel Density Estimation (KDE) curves visualise the density of prompt-level win rates (the proportion of $N=50$ trials in which a topic is mentioned for a given prompt). A concentration at $x=0$ indicates that the model is predisposed to omit the topic across the prompt set, while a concentration at $x=1$ reflects a systematic tendency to include it.} 
    \label{fig:density}
\end{figure*}

Confirming the effect observed in \autoref{sec:homo_res}, \autoref{fig:density} shows how leading models generally follow the frequency patterns observed in smaller post-trained models, often increasing the promotions or suppression of a topic. A notable exception is \textit{Carbon Pricing}, which leading models promote more actively than previous generations and \textit{Nuclear Power}, which is much more frequently present in leading outputs. Otherwise, leading models exhibit either an even higher concentration at zero incidence (no mention), or push popular topics toward systematic representation (1, always mentioned). 

CoT does mitigate some of the extremes of leading models, such as the systematic promotion of \textit{Renewable Energy} or  \textit{Energy Efficiency} and some of the erasure of \textit{Alternative Economic Systems} and \textit{Socio-Technical Approaches}. However, its impact remains limited. Notably, the distributions for the latter two topics remain heavily concentrated around zero. It also seems to introduce the suppression of other topics, such as \textit{Clean Transportation}.

\textit{ClimateGPT} seem to, in general, preserves more of the distributional trends observed in pre-trained models, although it still shows some promoting or suppressing tendency, especially concerning \textit{Alternative Economic System}. 

\subsection{Assessing Generalisation}

\begin{figure}[hbt!]
    \centering
    \includegraphics[width=0.7\columnwidth]{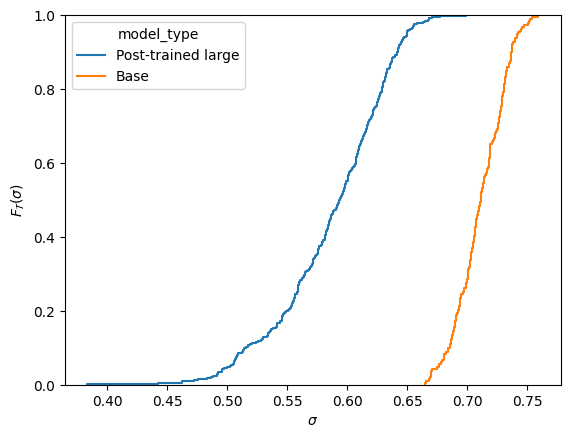}
    \caption{\textbf{Poverty and Homelessness ECDF}}
    \label{fig:poverty_ecdf}
\end{figure}

\begin{figure}[hbt!]
    \centering
    \includegraphics[width=\columnwidth]{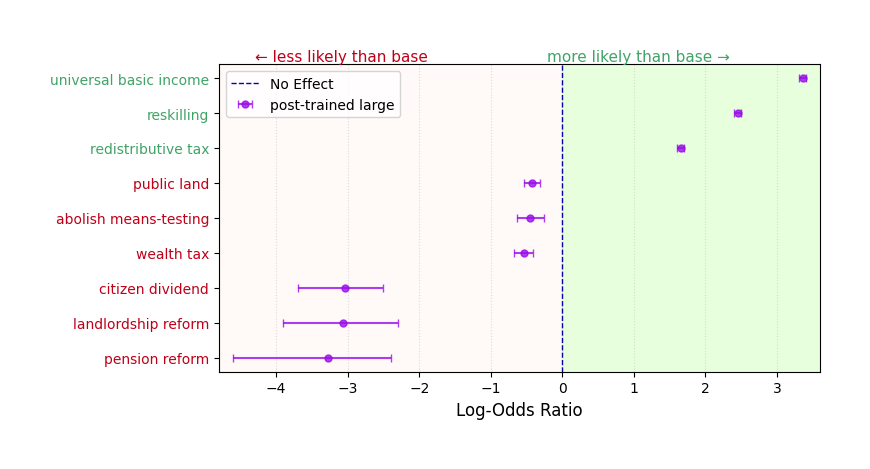}
    \caption{\textbf{Poverty and Homelessness Credible Intervals}}
    \label{fig:poverty_glmm}
\end{figure}

Results on the poverty and homelessness use-case show the same pattern as for the climate topic when comparing base vs. large post-trained models, as shown in \autoref{fig:poverty_ecdf} and \autoref{fig:poverty_glmm}. Alignment reduces semantic keyword spread (0.71 $\pm$ 0.02 $\rightarrow$ 0.59 $\pm$ 0.05) and systematically promotes topics such as reskilling (log-OR = 2.4) and universal basic income (log-OR = 3.3), whilst suppressing others, such as landlordship reform (log-OR = -3.1) and pension reform (log-OR = -3.3). The complete list of effects for all 64
topics is provided in \autoref{fig:glmm_full_poverty}.

\section{Discussion}

\subsection{The Consequences of Homogenisation}
Homogenisation metrics confirm that post-training reduces diversity and that interventions such as specialised fine-tuning or CoT prompting can partially recover it. However, these aggregate measures alone cannot reveal what is actually being lost: which topics are suppressed and promoted, and what this means for climate discourse. By quantifying shifts in topic saliency post-training, we can discern clear 'winners and losers' in the model’s topical hierarchy. While topics like alternative energy and energy efficiency become near-systematic, appearing up to almost 50 times more frequently, socio-political and alternative economic keywords rarely appear after post-training. Interestingly, while leading models mostly follow the promotional patterns of smaller open-source models, they sometimes exhibit a more polarised narrowing: they advocate more aggressively for carbon pricing and nuclear energy while more sharply pruning alternative economic or sociopolitical frameworks. 

This characterisation illuminates the societal consequences of diversity loss. It reveals a coherent directional homogenisation at an ecosystem level: across both small and leading models, energy and infrastructure solutions are prioritised over alternative socio-political frameworks or radical economic changes. Such a pattern mirrors an existing 'technical-fix' bias which has already been identified within the climate discourse literature. As \citet{obrien15degCTargetPossible2018} argues in her work, while there are three potential spheres of transformation, much of the current discourse focuses on the practical (technical solutions) or personal sphere, at the expense of the political spheres (social and systemic change).  This narrowness fundamentally limits our collective ability to reach climate targets, and the authors argue that change can only be meaningfully leveraged when all three spheres are being acted upon. 

Thus, by narrowing the reflected solutions set, LLMs might contribute to the marginalisation of critical dimensions of transformation and collapse the interpretation of what constitutes a `good' climate solution, obscuring the fact that social changes are often more rapid and cost-effective for emission reduction than technological fixes alone \cite{shuklaClimateChange20222022}. In this light, LLMs may actually flatten the necessary nuanced and `wicked' reality of the climate crisis, potentially narrowing the public’s perception of the possible, resulting policy agendas, and ultimately, the range of attempted mitigations.

\subsection{Assessing Mitigations Approaches}

While homogenisation metrics suggest that both CoT prompting and specialised fine-tuning mitigate diversity loss, quantifying shifts in topic saliency allows for a more nuanced interpretation of this result. It reveals that while neither approach fully avoids topic suppression, ClimateGPT maintains a frequency profile most comparable to base models. This finding suggests that maintaining diverse perspectives on, among others, climate solutions may require adopting smaller, specialised models, rather than relying on a small number of centrally-aligned, generalist models. Nonetheless, the suppressive tendencies of such models still necessitate careful monitoring. By showing that pluralism requires specialisation, we provide empirical support for a decentralised LLM ecosystem, one comprising diverse, specialised models rather than globally-distributed generalist systems optimised for universal alignment \cite{lavazzaGenerativeAIKnowledge2026, peterDecentralisingLLMAlignment2025, varshneyDecolonialAIAlignment2024}. Furthermore, our approach allows to qualify the pluralisms of these specialised models. 

\subsection{Generalisation}
Extending our evaluation to a new area confirms that LLMs restrict solution diversity across multiple domains. Like climate change, there is no single silver bullet for solving poverty and homelessness. Focusing solely on demand-side policies, like universal basic income, leaves underlying supply-side causes unaddressed, such as predatory evictions and profit-maximising practices \cite{hilberHousingPolicyAffordable, aminSpatiotemporalAnalysisEvictions2026}. When models under-report systemic solutions like landlordship reform, they narrow the range of actionable options presented to decision-makers, ultimately impairing effective public policy design. Although fully characterising topic saliency shifts in this domain requires further study, this experiment confirms the broad relevance of our methodology and shows how systemic homogenisation undermines a wide range of multi-faceted problem solving .

\section{Conclusion}
In this work, we provide empirical evidence that quantifying shifts in topic saliency is essential to delineate the potential societal effect of homogenisation and whether proposed mitigations address these risks. We introduce a novel methodology to identify these shifts, and apply it to climate change discourse across major LLM families. Our analysis reveals that post-training substantially narrows proposed solutions: renewable energy becomes almost 50 times more likely to appear, whilst nuclear, socio-political, and alternative economic solutions credibly decrease in salience. This flattening undermines the nuanced responses necessary for addressing complex societal challenges.

We also showed that specialised fine-tuning can partially mitigate these effects. This provides empirical evidence that decentralising AI systems could help counter the worsening of generative monoculture. When generalist models are used to distribute knowledge and perspectives on complex social issues, they risk collapsing the epistemic range of options deemed feasible and centralising agenda-setting power. Our method provides a tool to identify and characterise this effect across use cases. We also demonstrated how this narrowing extends beyond climate change to other wicked problems such as poverty and homelessness. 

In the future, our approach could be applied to study the narrowing of LLM representations on other multifaceted social challenges, such as global financial stability \cite{mainelli2008wicked} or the erosion of democracy \cite{theisPoliticalScienceCivic2016}. It could also be extended to study impacts beyond post-training, such as the biasing effects of Retrieval-Augmented Generation \cite{daiBiasUnfairnessInformation2024} or the narrowing caused by model collapse \cite{schaefferPositionModelCollapse2025}.

\section{Limitations}

Our proposed methodology has several limitations, which mostly stem from an effort not to overcomplicate it. In particular, we acknowledge that in production environments, LLMs' chat interfaces or agents are frequently deployed within Retrieval-Augmented Generation (RAG) pipelines, which might change the saliency of mentioned topics, depending on their presence in the retrieved set. However, the generator's inherent biases remain a critical point of failure \cite{wangAstuteRAGOvercoming2025}, and it is thus essential to understand the model's internal tendency to suppress or emphasise certain topics, as this can persist even when contrary information is present in the retrieved context. 

Additionally, our prompts are not conditioned by pre-existing information about a user, and consequently, they may not encompass the full breadth of colloquial or persona-based interactions typical of consumer chatbots. Using a standardised, technical query format ensures that the observed homogenisation is a reflection of the models' underlying conceptual priors rather than a reaction to specific sociolinguistic cues or persona-driven steerability, but might reduce the diversity of elicited topics. 

Furthermore, while we designed our methodology to be agnostic to the specific caused of homogenisation, we only demonstrated it's effectiveness with post-training alignment. The same controlled-comparison framework could apply to other interventions, such as the diversity loss of model collapse, where the baseline could become earlier models. More work is necessary to assess the impacts of other architectural or training decisions besides post-training.

Ultimately, our study thus serves as a necessary baseline for the default state of these models and can serve as a basis for understanding the behaviour of more complex, integrated AI systems. Future work should expand this mapping to include diverse user personas and investigate how the introduction of external knowledge via RAG interacts with the internalised homogenisation documented here.

\section{Ethical Considerations}

While this paper describes the risks of homogenisation, it is important to acknowledge that outputs' diversity for diversity's sake should not be the goal either. For example, while post-training reduces the plurality of reproduced viewpoints, maximising for diversity inherently risks reintroducing the harmful or toxic perspectives that post-training seeks to reduce. This trade-off underscores the necessity of identifying which topics are suppressed during homogenisation, as some exclusions may indeed be ethically desirable. While we demonstrate that certain excluded topics are important to climate discourse, the objective is not the blanket elimination of safety filters, but rather ensuring that no \textit{valuable} topic is erased. 

Optimising for this boundary, however, immediately forces the contentious question of who decides what \textit{valuable} means, and conversely, what constitutes \textit{toxic} or \textit{harmful}. Grounding these investigations in specific contexts allows anchoring these value-laden concepts, for example, defining the terms differently for climate change or children's education, but it does not resolve the underlying question of epistemic authority.

Furthermore, the solutions we propose, such as decentralising or specialising post-training, remain interventions confined to the model level. However, broader systemic, institutional, and regulatory efforts are required to effectively mitigate the risks of generative monoculture and knowledge collapse.

\section*{Acknowledgments}
This work was supported by the Engineering and Physical Sciences Research Council, grant number EP/Y009800/1, Responsible AI UK.
\bibliography{custom}

\appendix

\section{Appendix}
\label{sec:appendix}
\subsection{Methodological Comparisons} \label{sec:detail_dis}
While our method is intended as a novel and complementary focus on LLM homogenisation measurement, it borrows much from pluralism research. Here, we therefore discuss how it diverges from both these domains. 

\subsubsection{Comparisons to Homogenisation}
LLM homogenisation is commonly assessed by sampling model responses to a variety of prompts, where a larger resulting output space signifies greater diversity \cite{havrillaSurveyingEffectsQuality2024}. A reduction in output diversity is consequently identified by a decrease in the coverage of this potential sampling space. However, estimating the potential output space requires a broad range of inputs, leading to a lack of topic-level differentiation in results. We aim to fill this gap in our paper, and are therefore not proposing an \textit{alternative} to existing homogenisation measurements, but rather a \textit{complementary} analysis at the topic level. 

Consequently, our proposed approach relies on pre-existing semantic-based homogenisation metrics, see \autoref{fig:sementicspread}. There are four common ways to quantify the diversity of outputs of a model: \textit{Lexical dispersion} serves as the most granular approach, assessing inherent vocabulary and word choice variety through n-gram analysis. While useful for identifying repetitive phrasing, it often fails to capture conceptual overlaps. \textit{Syntactic dispersion} offers a higher-level view by evaluating variations in grammatical structures and sentence constructions. Moving further into the conceptual realm, \textit{logical diversity} utilises Natural Language Inference (NLI) models to focus on the variety of propositions or opinions conveyed. While this is a common measure for pluralism (see \autoref{tab:pluralism}), it is not adapted to identifying thematic similarity. Finally, \textit{semantic dispersion} moves beyond surface features to quantify diversity based on the text's underlying embeddings, used as a proxy for meaning. 

While we prioritise a semantic measure to capture the thematic clustering central to our analysis, we apply this metric to extracted keywords rather than full-text outputs. This approach implicitly accounts for lexical diversity; because identical terms share the same vector representation, their literal repetition naturally reduces the pairwise semantic distance, reflecting both conceptual and vocabulary-based narrowing.

\autoref{tab:dispersion} provides an overview of these dispersion-based methods, their associated metrics, and references.

\begin{table*}[h!]
\centering
\begin{tabular}{lp{3cm}p{7cm}}
\hline
\textbf{Type} & \textbf{Target} & \textbf{Metrics} \\
\hline
Lexical Dispersion & n-gram & Distinct \cite{shiThoroughExaminationDecoding2024, kirkUnderstandingEffectsRLHF2024, guoBenchmarkingLinguisticDiversity2024, guoCuriousDeclineLinguistic2024, zhangGenerativeAIMeets2025}, Entropy \cite{dhamalaAnalysisEffectsDecoding2023}, Jaccard index \cite{omahonyAttributingModeCollapse2024, wangLargeLanguageModels2025}, Normalised Compression \cite{omahonyAttributingModeCollapse2024}, Token-Type Ratio \cite{guoCuriousDeclineLinguistic2024, guoBenchmarkingLinguisticDiversity2024, agarwalAISuggestionsHomogenize2025}, Bleu \cite{baoDecodingMattersAddressing2024}, Self-Bleu \cite{omahonyAttributingModeCollapse2024, guoCuriousDeclineLinguistic2024, holtzmanCuriousCaseNeural2019, chenSpiralSilenceHow2024}, Rouge-L \cite{padmakumarDoesWritingLanguage2024} \\
\hline
Syntactic Dispersion & Morphosyntactic Pattern & Entropy \cite{luoDivergeNotDiverge2024} \\
\cline{2-3}
& Syntactic Graph & DivSyn \cite{guoBenchmarkingLinguisticDiversity2024, guoCuriousDeclineLinguistic2024} \\
\hline
Semantic Dispersion & Word Embeddings (Glove) & Pairwise Cosine Distance \cite{koivistoBestHumansStill2023, zhangGenerativeAIMeets2025} \\
\cline{2-3}
& Sentence Embeddings (Transformer) & Pairwise Cosine Distance \cite{omahonyAttributingModeCollapse2024, kirkUnderstandingEffectsRLHF2024, guoBenchmarkingLinguisticDiversity2024, guoCuriousDeclineLinguistic2024, wuGenerativeMonocultureLarge2024, agarwalAISuggestionsHomogenize2025, padmakumarDoesWritingLanguage2024, wangLargeLanguageModels2025, jiangArtificialHivemindOpenEnded2025}, Covariance Matrix Determinant \cite{wangLargeLanguageModels2025}, Vendi Score \cite{wangLargeLanguageModels2025}, Cosine Distance to 'Average' Group Embedding \cite{andersonHomogenizationEffectsLarge2024, doshiGenerativeAIEnhances2024}, Pairwise Dot-Product \cite{dellacquaNavigatingJaggedTechnological2023}, Pairwise Jaccard index \cite{wuGenerativeMonocultureLarge2024}, Pairwise Fingerprint Similarity \cite{wuGenerativeMonocultureLarge2024}\\
\cline{2-3}
& Other Text Embedding & tf-idf + Pairwise Cosine Distance \cite{zhangGenerativeAIMeets2025}, LIWC + Pairwise Cosine Distance \cite{alveroLargeLanguageModels2024}, CIELAB + Pairwise Perceptual Similarity \cite{murthyOneFishTwo2025}\\
\hline
'Logic' Dispersion & NLI Prediction & Contradiction Frequency \cite{kirkUnderstandingEffectsRLHF2024, fengModularPluralismPluralistic2024} \\
\hline
\end{tabular}
\caption{Summary of Existing Homogenisation Metrics. NLI refers to Natural Language Inference}
\label{tab:dispersion}
\end{table*}  

\subsubsection{Comparisons to Pluralism}

LLM Pluralism research, in contrast, looks to characterise what models do or do not output. Much of this literature evaluates what \citet{sorensenPositionRoadmapPluralistic2024} categorises as \textit{distributional pluralism}, where researchers compare value-laden opinion distributions in model outputs to those of human populations. This is typically measured by calculating the distance between a model's probability weights on specific tokens and the distribution of a human survey population \cite{durmusMeasuringRepresentationSubjective2023, sorensenPositionRoadmapPluralistic2024, santurkarWhoseOpinionsLanguage2023, atariWhichHumans2023, wrightLLMTropesRevealing2024, rozadoPoliticalPreferencesLLMs2024}. 

However, \citet{rottgerPoliticalCompassSpinning2024} demonstrates that these methodologies often lack robustness due to extreme format sensitivity, where models yield vastly different results depending on prompt wording or when forced-choice surveys are replaced with free-text open-ended queries. They argue that this instability stems in part from an attempt to attribute discrete, static opinions to LLMs, when they are best understood as a superposition of personas. Furthermore, they suggest increasing robustness through free-text evaluations within a well-scoped task. In a subsequent study, they built on this insight to explore the distributions of models' reflected stances on political topics, demonstrating consistent and convergent bias across providers \cite{rottgerIssueBenchMillionsRealistic2026}. 

While this latter paper is motivated by the same concern as ours is, namely that LLMs often emphasise specific perspectives and ideas at the expense of others, our methodologies diverge in scope. \citet{rottgerIssueBenchMillionsRealistic2026} investigates the coverage of perceptions of LLMs, or their bias over a static set of 216 issues. In contrast, this work focuses on the thematic breadth to understand the coverage or diversity of issues which can be elicited in the first place. 

Our framing of plurality is therefore closer to what \citet{sorensenPositionRoadmapPluralistic2024} qualifies as \textit{Overton pluralism}, or the coverage of a spectrum of reasonable responses. Existing studies on \textit{Overton pluralism} quantify how much of an identified set of values is covered by a model's outputs \cite{fengModularPluralismPluralistic2024, shettyVITALNewDataset2025}, or to what extent a human cohort agrees with a model’s statement \cite{poole-dayan2026benchmarking}. Yet, they do not discuss which specific topics are included or excluded from the model's `Overton window' or the implications of such omissions.

Our work thus occupies a middle ground between Overton and distributional pluralism by shifting the unit of analysis from surface-level tokens to topic saliency shifts. We are concerned with both the breadth of the Overton window, the set of topics mentioned in relation to an issue, and the probability mass assigned to each of these topics. In particular, we measure how post-training or diversity mitigation techniques influence this likelihood. By evaluating the `distributional' aspect of the `Overton' set, we can identify credible increases or decreases in the prominence of specific themes and their implications, filling this research gap. 

\autoref{tab:pluralism} provides an overview of existing measures of \textit{distributional} and \textit{Overton} pluralism.

\begin{table*}[h]
\centering
\begin{tabular}{lp{3cm}p{7cm}}
\hline
\textbf{Type} & \textbf{Target} & \textbf{Metrics} \\ 
\hline
Distributional
& Human vs. token distributions 
& Jensen-Shannon \cite{pavlovicUnderstandingEffectTemperature2024, durmusMeasuringRepresentationSubjective2023, sorensenPositionRoadmapPluralistic2024, fengModularPluralismPluralistic2024, leeCanLargeLanguage2023, shettyVITALNewDataset2025}, Wasserstein \cite{santurkarWhoseOpinionsLanguage2023} \\ 
\cline{2-3}
& Topic stance distribution per models 
& Jensen-Shannon \cite{rottgerIssueBenchMillionsRealistic2026}\\ 
\hline
Overton 
& LLM output vs. value list
& NLI \cite{fengModularPluralismPluralistic2024, shettyVITALNewDataset2025}, LLM-as-a-Judge Win Rate \cite{fengModularPluralismPluralistic2024, shettyVITALNewDataset2025} \\ 
\cline{2-3}
& Human agreement 
& Average Coverage (proportion of human viewpoints represented within outputs) \cite{poole-dayan2026benchmarking} \\ 
\hline
\end{tabular}
\caption{Summary of Existing Distributional and Overton Pluralism Metrics}
\label{tab:pluralism}
\end{table*}

\subsection{Methodological Details}
\subsubsection{Vanilla Prompting}
\label{sec:appendix-prompts}
To ensure a controlled comparison between pre-trained and post-trained/instruction-tuned models, we adapted our prompting strategy to align with each model's training objective. For pre-trained models, which are trained for text completion, we used the following template: ``\textit{If I had to describe the \{adj\} concrete \{noun\} to \{entity\} in a sentence, I would write the following:}''. In contrast, instruction-tuned models were prompted using this instruction format: ``\textit{In a sentence, describe the \{adj\} concrete \{noun\} to \{entity\}.}''. For the OLMo model, we further wrapped the post-trained prompt in the required \texttt{<|im\_start|>} and \texttt{<|im\_end|>} tokens to match its instruction-tuning format. We limited the number of generated tokens to 100.  The specific lexical components used to populate these templates are listed in \autoref{tab:lexical-components-climate} for the climate use-case and \autoref{tab:lexical-components-poverty} for the one on poverty and homelessness. We took the Cartesian product of these sets ($4 \times 4 \times 4$) to generate 64 unique prompt instances, allowing us to account for the influence of specific phrasing.

\begin{table*}
  \centering
  \begin{tabular}{l l l}
    \hline
    \textbf{Adjectives} & \textbf{Nouns} & \textbf{Entities Climate}\\ 
    \hline
    best & approach & survive climate change  \\
    most efficient & action & climate adaptation \\
    most promising & pathway & resolve the climate crisis \\
    recommended & solution & achieve long-term climate sustainability \\\hline
  \end{tabular}
  \caption{Lexical components used in climate prompt generation.}
    \label{tab:lexical-components-climate}
\end{table*}

\begin{table*}
  \centering
  \begin{tabular}{l l l}
    \hline
    \textbf{Adjectives} & \textbf{Nouns} & \textbf{Entities Poverty} \\ 
    \hline
    best & approach & solve poverty and homelessness  \\
    most efficient & action & ensure everyone has basic financial security and a stable place to live \\
    most promising & pathway & guarantee sufficient income and reliable shelter for all \\
    recommended & solution & overcome severe economic hardship and residential instability \\\hline
  \end{tabular}
  \caption{Lexical components used in poverty prompt generation.}
    \label{tab:lexical-components-poverty}
\end{table*}

\subsubsection{CoT procedure} \label{sec:cot}
We designed the Chain-of-Thought (CoT) procedure by adapting the diversity-enhancing approach of \citet{meinckePromptingDiverseIdeas2024}, recontextualising it from product design to climate solutions. Our implementation delivers instructions sequentially for two primary reasons: first, to simulate the multi-turn interactions typical of conversational chatbots; and second, to isolate the final response for direct comparison with vanilla prompting. By keeping the final instruction identical across both methods, we ensure that subsequent analyses of topic saliency remain methodologically consistent.

The sequential procedure follows this structure, using the variable \texttt{base} (e.g. "the best concrete action to resolve the climate crisis"):
\begin{itemize}
\item Instruction 1: "Generate a list of 15 ideas on \{\texttt{base}\} (short titles only)."
\item Instruction 2: "Review the list and determine whether the ideas are distinct and bold; modify them as needed to ensure maximum variance. No two ideas should be the same."
\item Instruction 3 (Final Output): "In a sentence, describe \{\texttt{base}\}."
\end{itemize}

See \autoref{tab:cot_example} for an example of a CoT completion. 
\begin{table*}[h!]
\centering
\renewcommand{\arraystretch}{1.5}
\begin{tabular}{p{1.5cm} p{2cm} p{10cm}}
\hline
\textbf{Turn \#} & \textbf{Role} & \textbf{Content} \\ \hline
1 & \textbf{User} & Generate a list of 15 ideas on the best concrete approach to achieve long term climate sustainability (short title only) \\ \hline
2 & \textbf{Assistant} & 1. Net-Zero Grid Electrification, 2. Deep Building Energy Retrofits, 3. Zero-Carbon Concrete Mixes, 4. Low-Carbon Cement Substitution, 5. Industrial Heat Decarbonization, 6. Methane Leak Detection \& Repair, 7. Citywide Heat Pumps Deployment, 8. Circular Construction Materials, 9. Low-Carbon Transportation Networks, 10. EV Charging \& Fleet Electrification, 11. Renewable Energy Storage Buildout, 12. Nature-Based Carbon Sequestration, 13. Climate-Resilient Water Infrastructure, 14. Urban Forest \& Cool Roof Programs, 15. Climate-Aligned Policy \& Carbon Pricing \\ \hline
3 & \textbf{User} & Go through the list and determine whether the ideas are different and bold, modify the ideas as needed to make them bolder and more different. No two ideas should be the same. This is important! \\ \hline
4 & \textbf{Assistant} & 1. Build a 100\% Renewable “Transnational Grid”, 2. Mandate All New Buildings to Be Net-Plus-Energy, 3. Replace Cement With Fully Alternative Binders, 4. Cap Cement Carbon Intensity to Ultra-Low Targets, 5. Electrify Steel Kilns and Cement Plants Nationwide, 6. Zero-Tolerance Methane Policy With Continuous Monitoring, 7. Deliver Heat-Pump Microgrids for Every Neighborhood, 8. Require True Circular Construction (Design for Disassembly), 9. Shift to Zero-Emission Freight Corridors by 2030, 10. Ban Diesel Fleets; Full Depot-to-Route Electrification, 11. Standardize Utility-Scale Storage as Core Grid Infrastructure, 12. Pay for Verified Soil Carbon Through Regenerative Farming at Scale, 13. Harden Water Systems for Extremes With Real-Time Control, 14. Scale “Cool City” Infrastructure: High-Albedo Streets Everywhere, 15. Enforce Carbon-Transparent Supply Chains With Legal Liability \\ \hline
5 & \textbf{User} & In a sentence, describe the best concrete approach to achieve long term climate sustainability. \\ \hline
6 & \textbf{Assistant} & The best concrete approach is to rapidly decarbonize buildings, grids, and industry with proven electrification plus low-carbon materials, while cutting methane, enforcing circular/efficient construction, and funding long-term measurement-backed resilience. \\ \hline
\end{tabular}
\caption{Example of Chain of Thought Output for the prompt ``best concrete approach to achieve long term climate sustainability''}
\label{tab:cot_example}
\end{table*}

\subsubsection{Keywords Extraction} \label{sec:topic-extraction}
Keyword extraction was performed using the \texttt{openai/gpt-5-mini} model on \texttt{openrouter}. To optimise for consistency and extraction quality, the model's reasoning effort was set to `minimal', and it was provided with a few-shot prompt designed for information retrieval within the climate adaptation domain. The prompt, shown below, explicitly instructs the model to identify specific actors, metrics, and actions while ensuring the extracted terms are relevant to the input question. Examples of outputs and extracted keywords can be found in \autoref{app:climate_keywords}. 

To validate the extraction, we manually annotated 195 completions and compared the keywords extracted by texttt{LLaMA 3.1 70B} against \texttt{gpt-5-mini}. Matches were identified by retrieving the closest fuzzy match via the Python \texttt{rapidfuzz} library, accepting pairs with a similarity score above 50. The resulting validation metrics are detailed in \autoref{tab:kw_validation}. Furthermore, we categorised the unextracted keywords to confirm that no specific topic was systematically missed or underestimated by our approach. At most 0.5\% of missed keywords belonged to any single category, indicating that omission errors were uniformly distributed and did not introduce systematic bias. 

\begin{table*}[ht]
\centering
\small 
\begin{tabular}{l l l} 
\hline
\textbf{Model} & \textbf{F1} & \textbf{Recall} \\
\hline
\texttt{LLaMA 3.1 70B} & 0.75 & 0.7 \\
\hline
\texttt{gpt-5-mini} & 0.84 & 0.94 \\
\hline
\end{tabular}
\caption{Fuzzy matching score against manual annotation of keywords extraction}
\label{tab:kw_validation}
\end{table*}

\begin{table*}[t] 
\begin{tcolorbox}[
    colback=gray!5, 
    colframe=gray!50, 
    title=Keyword Extraction Prompt, 
    fonttitle=\bfseries,
    fontupper=\small\ttfamily,
]
Role: You are an expert Linguistic Analyst specialising in Information Retrieval in the context of Climate Adaptation. \\
Task: Extract highly specific key terms from EACH of the provided Responses, using the Question to determine relevance. \\
\\
Instructions:
\begin{itemize}
    \item Length: Each term must be between 1 and 4 words long.
    \item Relevance: Only extract terms that directly address or explain the core subject of the Question. Do not extract terms that are only found in the question itself. If no relevant terms exist in the Answer, return an empty list.
\end{itemize}
Examples:
\begin{itemize}
    \item \textbf{Example 1:} \\
    Question: In a sentence, describe the best concrete action to survive climate change. \\
    Response: To survive climate change, one of the most effective concrete actions individuals can take is to drastically reduce their carbon footprint by transitioning to a plant-based diet, minimizing air travel, and using renewable energy sources such as solar or wind power for their homes and transportation. \\
    Keywords and Rational: \\
    - 'individuals' -> Climate adaptation actors. \\
    - 'carbon footprint' -> Climate adaptation metrics. \\
    - 'plant-based diet', 'minimizing air travel', 'renewable energy', 'solar power', 'wind power' -> Climate adaptation actions.
    
    \item \textbf{Example 2:} \\
    Question: In a sentence, describe the best concrete action to resolve the climate crisis. \\
    Response: To resolve the climate crisis, the best concrete action is to rapidly transition to 100\% renewable energy worldwide by investing in solar, wind, and other clean energy technologies, improving energy efficiency, and implementing policies like carbon pricing and green infrastructure development, as outlined in reports by organizations such as IRENA and the IPCC. \\
    Keywords and Rational: \\
    - 'IRENA', 'IPCC' -> Climate adaptation actors. \\
    - 'renewable energy', 'clean energy technologies', 'improving energy efficiency', 'implementing policies', 'carbon pricing', 'green infrastructure development' -> Climate adaptation actions.
\end{itemize}
Input: \\
Question: \{question\} \\
Responses: \{template\}
\end{tcolorbox}
\end{table*}

\subsubsection{Keywords Classification} \label{sec:topic-classification}
To ground the climate keywords classification in human annotations, we started by clustering and labelling a reference set of 12,000 keywords randomly sampled from keywords extracted from our dataset. We then manually curated the clusters into distinct topics. We then used this reference to classify the remaining keywords into these topics. The same procedure was repeated on 1,314 poverty and homelessness keywords.

\paragraph{Embeddings} All extracted keywords were transformed into high-dimensional vectors using the OpenAI \texttt{text-368-embedding-3-small} model. This provided the semantic foundation for all subsequent clustering and classification steps.
\paragraph{Clustering} First, we performed an exploratory grouping of the 12,000 reference keywords using a density-based clustering pipeline. To mitigate the curse of dimensionality, we first reduced the embeddings to 20 dimensions using UMAP ($n\_neighbors=15$) \cite{mcinnesUMAPUniformManifold2018}. We then applied HDBSCAN using Euclidean distance ( $min\_size=50$ and $min\_samples$=15) \citep{campelloDensityBasedClusteringBased2013}.

\paragraph{Topic Attribution} The resulting clusters were qualitatively reviewed; clusters were merged or split where necessary to ensure conceptual consistency. Each refined cluster was then manually assigned a descriptive topic name to serve as a ground-truth label for the reference set.
\paragraph{K-Nearest Neighbours (KNN) Classification} The remaining keywords were classified using a K-nearest neighbour (KNN) approach \cite{hodges1951discriminatory}. To strengthen the semantic signal, reference embeddings were generated using a structured template: `Topic: \{topic\_name\}, Keywords: \{keywords\}'

For each new keyword, the 20 nearest neighbours were retrieved. A topic was assigned if it represented at least 30\% of the neighbours (a plurality threshold). This process successfully categorised 97.5\% of the keyword corpus. The negligible volume of unlabeled keywords suggests that our initial 12,000-word reference set provided sufficient coverage of the topic space; consequently, unlabeled keywords were discarded.

\paragraph{Validation} To validate the KNN classification, we conducted a manual audit on a stratified sample of the newly labelled keywords (approximately 10 examples per topic). During reannotation, the KNN-assigned labels were hidden to prevent bias. The classifier achieved 88\% accuracy. A manual error analysis revealed that most discrepancies occurred in ambiguous cases where the KNN's choice was still contextually defensible. For instance, the keyword `buy less plastic' was manually labelled as `individual action', while the KNN categorised it as `degrowth'. Therefore, we deemed the KNN performance sufficient for the remainder of the analysis. 

\subsubsection{Budget}
Large models output generation, keywords extraction and keywords embeddings were all computed using OpenRouter's infrastructure for less than 50£ in total. 

\subsection{Models}
\label{sec:appendix-models}
 Models with parameters up to $13\text{B}$ were executed on local infrastructure \cite{kingscollegelondone-researchteamKingsComputationalResearch2022} using the Hugging Face Transformers library \footnote{https://huggingface.co/docs/transformers/index} on NVIDIA GPUs. For larger models, inference was performed via the OpenRouter API \footnote{https://openrouter.ai}. To avoid artificial repetition of output across multiple trials due to provider-side caching, a unique \texttt{request\_id} was injected into the system prompt header for every API call. A detailed list of the models used in this work can be found in \autoref{tab:models-list}.

 \begin{table*}[ht]
\centering
\small
\begin{tabular}{l l l l l}
\hline
\textbf{Model Identifier} & \textbf{Category} & \textbf{Provider} & \textbf{Source} \\
\hline
\hline
\textit{Google (Gemma)} & &\\
Gemma-2-9B & Base & Local GPU & \cite{gemma_2024}\\
Gemma-2-9B-IT & Post-Trained & Local GPU  & \cite{gemma_2024}\\
Gemma-3-12B-PT & Base & Local GPU  & \cite{gemma_2025}\\
Gemma-3-12B-IT & Post-Trained  & Local GPU & \cite{gemma_2025} \\
Gemma-2-27B-IT & Post-Trained Large & OpenRouter & \cite{gemma_2024} \\
\hline
\textit{Meta (Llama)} & &\\
Llama-3.1-8B & Base & Local GPU & \cite{grattafiori2024llama3herdmodels} \\
Llama-3.1-8B-Instruct & Post-Trained & Local GPU  & \cite{grattafiori2024llama3herdmodels} \\
Llama-3.3-70B-Instruct & Post-Trained Large & OpenRouter  & \cite{grattafiori2024llama3herdmodels}\\
Llama-4-Maverick & Leading & OpenRouter & \cite{meta2024llama4} \\
\hline
\textit{Mistral AI} & &\\
Mistral-7B-v0.3 & Base & Local GPU & \cite{jiang2023mistral7b} \\
Mistral-7B-Instruct-v0.3 & Post-Trained & Local GPU & \cite{jiang2023mistral7b}\\
Mistral-Medium-3.1 & Post-Trained Large & OpenRouter & \cite{MediumNewLarge}\\
\hline
\textit{Allen AI (OLMo)} & &\\
Olmo-3-1025-7B & Base & Local GPU & \cite{olmo2026olmo3}\\
OLMo-3-7B-Instruct & Post-Trained & Local GPU & \cite{olmo2026olmo3}\\
OLMo-3.1-32B-Instruct & Post-Trained Large & OpenRouter & \cite{olmo2026olmo3}\\
\hline
\textit{Qwen} & &\\
Qwen2.5-7B & Base & Local GPU  & \cite{qwen2.5}\\
Qwen2.5-7B-Instruct & Post-Trained & Local GPU  & \cite{qwen2.5}\\
Qwen-2.5-72b-instruct & Post-Trained Large  & OpenRouter  & \cite{qwen2.5}\\
Qwen3-8B-Base & Base & Local GPU & \cite{qwen3technicalreport}\\
Qwen3-8B & Post-Trained & Local GPU & \cite{qwen3technicalreport}\\
Qwen3-next-80b-a3b-instruct & Post-Trained Large  & OpenRouter & \cite{qwen3technicalreport}\\
\hline
\textit{Swiss AI (Apertus)} & &\\
Apertus-8B-2509 & Base & Local GPU & \cite{swissai2025apertus}\\
Apertus-8B-Instruct-2509 & Post-Trained & Local GPU & \cite{swissai2025apertus}\\
Apertus-70B-Instruct & Post-Trained Large & OpenRouter  & \cite{swissai2025apertus}\\
\hline
\textit{OpenAI} & &\\
GPT-5-Nano & Leading & OpenRouter  & \cite{singh2026openaigpt5card}\\
\hline
\textit{Anthropic} & &\\
Claude-Haiku-4.5 & Leading & OpenRouter  & \cite{IntroducingClaudeHaiku}\\
\hline
\textit{xAI} & &\\
Grok-4.1-Fast & Leading & OpenRouter  & \cite{Grok41Fast}\\
\hline
\textit{Endowment for Climate Intelligence} & &\\
ClimateGPT-13B & ClimateGPT & Local GPU  & \cite{thulkeClimateGPTAISynthesizing2024}\\
\hline
\hline
\end{tabular}
\caption{The suite of LLMs evaluated in this study, by family and category used in the paper, citation and license.}
\label{tab:models-list}
\end{table*}

\subsection{Examples} \label{app:climate_keywords}
The lists of all topics and three representative, frequent keywords can be found in table \autoref{tab:climate_keywords} for the climate topic and in \autoref{tab:poverty_keywords} for the poverty and homelessness topic. Examples of answers of both base and post-trained models and extracted keywords can be found in \autoref{tab:output_examples}. 
\begin{table*}[ht]
\centering
\small 
\begin{tabular}{ll} 
\hline
\textbf{Topic Name} & \textbf{Keywords} \\
\hline
alternative economic system & green finance, diversifying economies, basic income \\
bio-fuel & green hydrogen, biochar, biofuels \\
biodiversity & ecosystem restoration, biodiversity, ecosystem-based adaptation \\
circular-economy & circular economy, circular reuse, regenerative economy \\
clean transportation & sustainable transportation, public transportation, low-carbon transportation \\
climate models & early warning systems, climate projections, data-driven planning \\
coast-protection and flood & sea walls, flood barriers, coastal protection \\
community-centric & community engagement, community preparedness, community-led planning \\
cost-benefit driven & cost-effective, benefit-cost ratios, cost-benefit analysis \\
decarbonisation & carbon capture and storage, carbon sequestration, natural carbon sinks \\
decentralisation & decentralized solutions, local adaptation, self-sufficiency \\
degrowth & reduce consumption, reduce energy consumption, consume less \\
design & climate-resilient design, sustainable design, design for disassembly \\
electrification & electrification, electrifying transportation, electrify buildings \\
emission reduction & reduce emissions, reduce carbon emissions, cutting methane \\
energy efficiency & energy efficiency, energy-efficient technologies, energy-efficient \\
extreme events & extreme weather, extreme heat, storms \\
food-system & sustainable land use, sustainable agriculture, plant-based diet \\
forest-centric & reforestation, mangroves, urban forests \\
funding-adaptation & investment, climate finance, public investment \\
global collaboration & international cooperation, ipcc, paris agreement \\
governance & equitable policies, policy changes, enforceable standards \\
greenhouse & greenhouse gas emissions, reduce greenhouse gas emissions, cut greenhouse gas emissions \\
human-nature equilibrium & nature-based solutions, human well-being, natural capital \\
individual action & carbon footprint, behavioral changes, behavioral change, lifestyle changes \\
industry-centric & industry, transforming industry, industrial emissions \\
infrastructure & resilient infrastructure, climate-resilient infrastructure, green infrastructure \\
knowledge-driven & education, local knowledge, research and development \\
managerial practices & continuous monitoring, proactive planning, adaptive management \\
material science & low-carbon concrete, permeable pavements, fly ash \\
nature management & restoring wetlands, soil health, conservation \\
net zero & net-zero emissions, net-zero by 2050, net-zero \\
nuclear & nuclear power, nuclear energy, nuclear fission \\
recycling & recycled aggregates, recycled materials, reduce waste \\
reduce fossil-fuel & phase out fossil fuels, replace fossil fuels, stop burning fossil fuels \\
renewable energy & renewable energy, solar power, wind power \\
risk-based approach & mitigation, reduce vulnerability, enhance resilience \\
smart-grid & grid modernization, smart grids, grid storage \\
socio-political approaches & collective action, systemic change, political will \\
state-led & national planning, government action, government intervention \\
sustainability & sustainable practices, sustainable development, durability \\
carbon pricing & carbon pricing, global carbon pricing, carbon tax \\
technology & green technologies, technological innovation, geoengineering \\
temperature-centric & urban heat island, limit warming 1.5°c, heat resilience \\
transition and transformation & just transition, rapid transition, equitable transition \\
urban planning & green roofs, urban green spaces, urban planning \\
vulnerable population centric & vulnerable communities, equity, climate justice \\
water management & water management, stormwater management, drought-resistant agriculture \\
\hline
\end{tabular}
\caption{Climate Change Topics and Most Frequent Associated Keywords}
\label{tab:climate_keywords}
\end{table*}

\begin{table*}[ht]
\centering
\small 
\begin{tabular}{ll}
\hline
\textbf{Topic Name} & \textbf{Keywords} \\
\hline
bank reform & abolish banking system, financial system reform, state-owned bank \\
build savings & financial literacy, emergency fund, personalized budget \\
carceral role & criminal justice reform, prevent criminalization, address criminal justice injustice \\
cash centric & targeted cash assistance, direct cash assistance, unconditional cash transfers \\
citizen dividend & social contract, social dividend, universal basic dividend \\
class concious & working class, social movement, grassroots movement \\
cost-benefit analyses & case management, strategic resource allocation, benefits \\
credit rating & credit-building, rebuilding credit, credit \\
debt reform & debt relief, debt management, reducing debt \\
decentralized & decentralized initiatives, democratic control, collective responsibility \\
dignity & human dignity, dignified home, dignity \\
economic reasoning & economic security, economic stability, economic resilience \\
education & education, education programs, accessible education \\
employment & stable employment, job creation, employment opportunities \\
end means-testing & no means-testing, no preconditions, non-means-tested financial floor \\
entrepreneurship & entrepreneurship, social entrepreneurship, entrepreneurial opportunities \\
eviction restrictions & tenant protections, eviction prevention, eviction prevention services \\
families & empower families, family stability, family planning \\
free market & local market needs, market volatility, free markets \\
universal basic income & universal basic income, guaranteed basic income, guaranteed minimum income \\
homeownership & homeownership, affordable homeownership, homeownership pathway \\
housing crisis & affordable housing supply, affordable housing construction, increase affordable housing supply \\
housing first & affordable housing initiatives, public housing, housing first \\
housing vouchers 	 & government-subsidized housing, subsidized housing, housing subsidies \\
immigration & emigration \& prevent displacement, emigration \& reduce displacement, emigration \& mobility \\
income supplementation & social safety nets, robust social safety nets, income support \\
intergenerational cycles & break cycles of poverty, break the cycle, break poverty cycles \\
job-training & job training, vocational training, job training and placement \\
labor market & labor protections, right to work, worker protections \\
labor productivity & earning potential, food production, means of production \\
landlordship reform & amend landlord laws, community-based landlords, abolition of landlordism \\
living-wage & living-wage jobs, living wage, guaranteed living-wage jobs \\
microfinance & microloans, financial inclusion, low-cost financial services \\
mix voucher & voucher programs, vouchers, cash or voucher \\
mixed-income communities & community engagement, community resources, mixed-income developments \\
mortgage & low-interest mortgages, affordable mortgages, interest-free loan \\
parenting support & childcare, childcare support, counseling \\
pension & state pension, pension inequality, universal basic pension \\
policy & policy reforms, policy changes, political interventions \\
preventive measures & addiction treatment, immediate relief, personal responsibility \\
post-capitalist economy & abolish capitalism, post-capitalist economy, avoid private capital accumulation \\
price indexation & indexed to inflation, cost of living index, fair pricing \\
private sector & public-private partnerships, high-demand sectors, cross-sector collaboration\\
public land & community land trusts, public ownership, access to land \\
racial disparities & systemic inequalities, racial disparities, discrimination \\
redistributive tax & progressive taxation, wealth redistribution, equitable taxation \\
rent-control & rent control, rent stabilization, rent control measures \\
retirement & financial independence, long-term financial planning, retirement income \\
security & financial security, foundational safety net, basic security \\
shelters & stable shelter, affordable shelter, secure shelter \\
social housing & social housing, universal affordable housing, social housing programs \\
standard-of-living & minimum standard of living, meet basic needs, stable living conditions \\
subsidies & wage subsidies, rent subsidies, government assistance \\
transportation & public transportation, accessible public services, free public transport \\
ubi & universal basic income (ubi), unemployment insurance, universal child benefit \\
universal healthcare & healthcare, accessible healthcare, universal healthcare \\
urban design & green infrastructure, right to the city, inclusive urban planning \\
vacant properties & vacant property repurposing, repurpose vacant properties, real estate crowdfunding \\
volunteering & non-profit organizations, local non-profit organizations, mutual aid \\
wealth tax & land-value taxes, closing tax loopholes, wealth tax \\
mental health & mental health care, mental health counseling, mental health resources \\
wraparound supports & mental health services, comprehensive support services, omprehensive social services \\
zoning inequities & equitable access, zoning reform, inclusionary zoning  \\
\hline
\end{tabular}
\caption{Poverty and Homelessness Topics and Most Frequent Associated Keywords}
\label{tab:poverty_keywords}
\end{table*}
\begin{table*}[ht]
\centering
\small 
\begin{tabular}{lp{3cm}p{7cm}} 
\textbf{Model Name} & \textbf{Keywords} & \textbf{Output} \\
\hline
Qwen/Qwen3-8B-Base (base) & 'vegan lifestyle', 'plant-based diet' & "If we continue to eat animals as we do now, the temperature will rise to 4 degrees Celsius or more by the end of this century, threatening our survival. Our health, economy, food supply, and social stability would all suffer if we keep using fossil fuels to produce meat. What a beautiful world it would be if we could all just consume plant-based foods instead of living in a chaotic atmosphere of thousands of species dying off." \\
\hline
Qwen/Qwen3-8B (post-trained) & 'renewable energy', 'fossil fuel use', 'mitigating global warming' & "The best concrete action to survive climate change is to transition to renewable energy sources such as solar and wind power, significantly reducing our reliance on fossil fuels and mitigating the impacts of global warming."  \\
\hline
Apertus-8B-2509 (base) & 'social cohesion', 'transparent decision making', 'knowledge production' & "Social cohesion based on radical, transparent and ethical decision making, as well as efficient organization of knowledge production and state planning" \\
\hline
Apertus-8B-Instruct-2509 (post-trained) & 'renewable energy sources', 'greenhouse gas emissions', 'green infrastructure' & "To address climate change, the best concrete solution is the immediate and rapid global transition to renewable energy sources, significantly reducing greenhouse gas emissions while promoting sustainable practices and green infrastructure development." \\
\hline
Olmo-3-1025-7B (base) & 'economic growth critique', 'UNEP GEO-6'' & "solve the economic system and you will solve the climate problem. And this might sound absurd, but in my opinion, all other solutions proposed so far, are nothing but short term political measures, which will leave us with another problem tomorrow. So we must finally agree that economic growth is bad for the climate. This is also the clear conclusion of the latest United Nations Environmental Program (UNEP) Global Environmental Outlook (GEO-6)." \\
\hline
Olmo-3-7B-Instruct (post-trained) & 'high-performance materials', 'optimized energy efficiency', 'reinforced concrete' & "A reinforced concrete structure built with high-performance materials and optimized for energy efficiency can provide excellent durability and resilience to withstand the impacts of climate change."\\
\hline
\end{tabular}
\caption{Examples of completions and extracted keywords for the prompt ``best concrete solution to survive climate change''}
\label{tab:output_examples}
\end{table*}

\subsection{Full GLMM Results}
\begin{figure*}[hbt!]
    \centering
    \includegraphics[width=\textwidth]{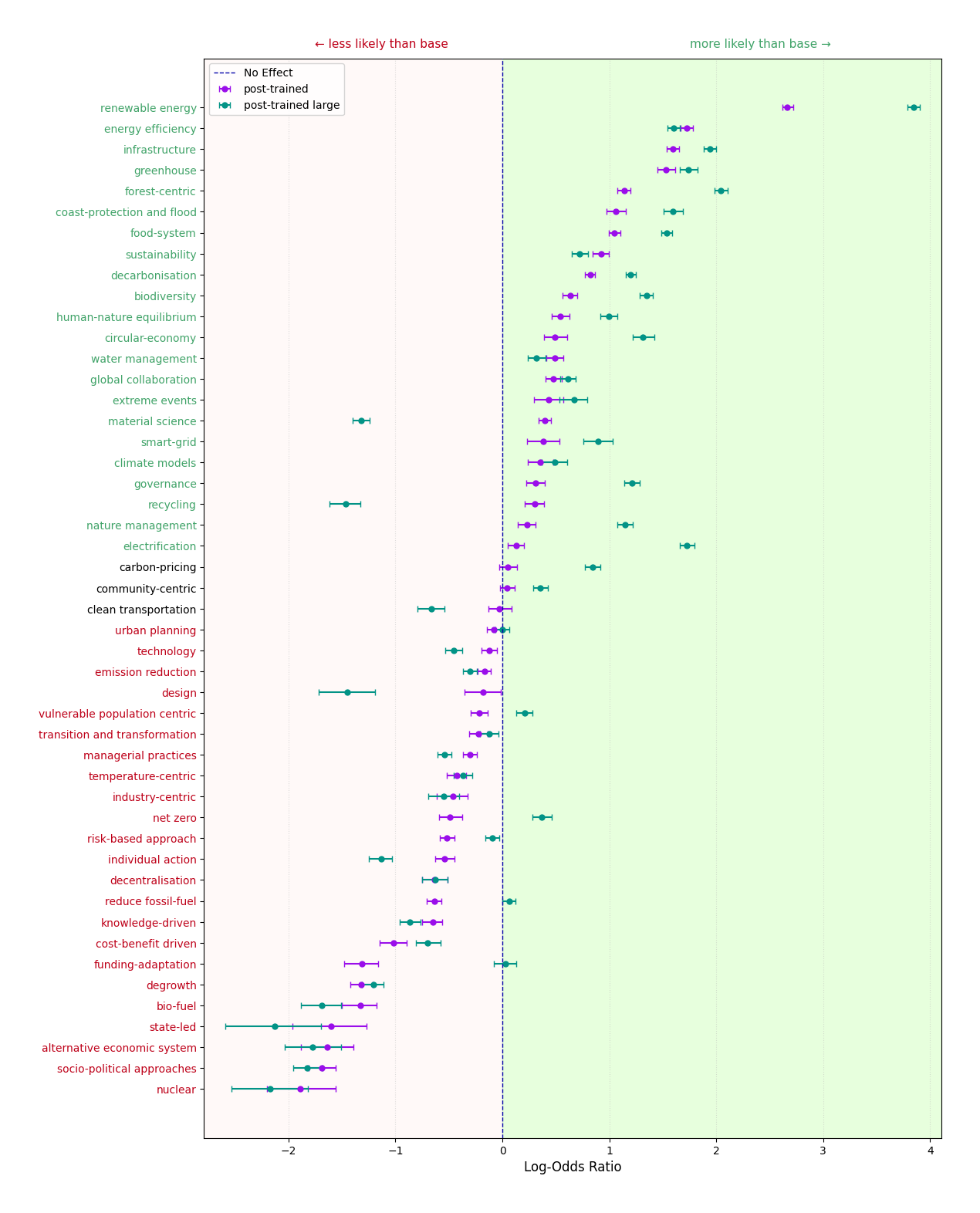}
    \caption{\textbf{Bayesian GLMM Credible Intervals for climate topic.} The plot displays the results for all 48 topics, ordered by the log-odds ratio of the post-trained effect. Intervals are shown for the post-trained (purple) and post-trained large (green) model effect relative to the pre-trained baseline. Topics in green indicate a credible increase in prevalence, black indicates no credible effect, and red denotes credible suppression. The log-odds ratio represents the shift in likelihood compared to the baseline; for example, a log-odds of -2.0 for \textit{Nuclear Energy} indicates it is exp(-2)~0.135 times as likely to appear, representing a roughly seven-fold decrease in prevalence.}
    \label{fig:glmm_full}
\end{figure*}

\begin{figure*}[hbt!]
    \centering
    \includegraphics[width=\textwidth]{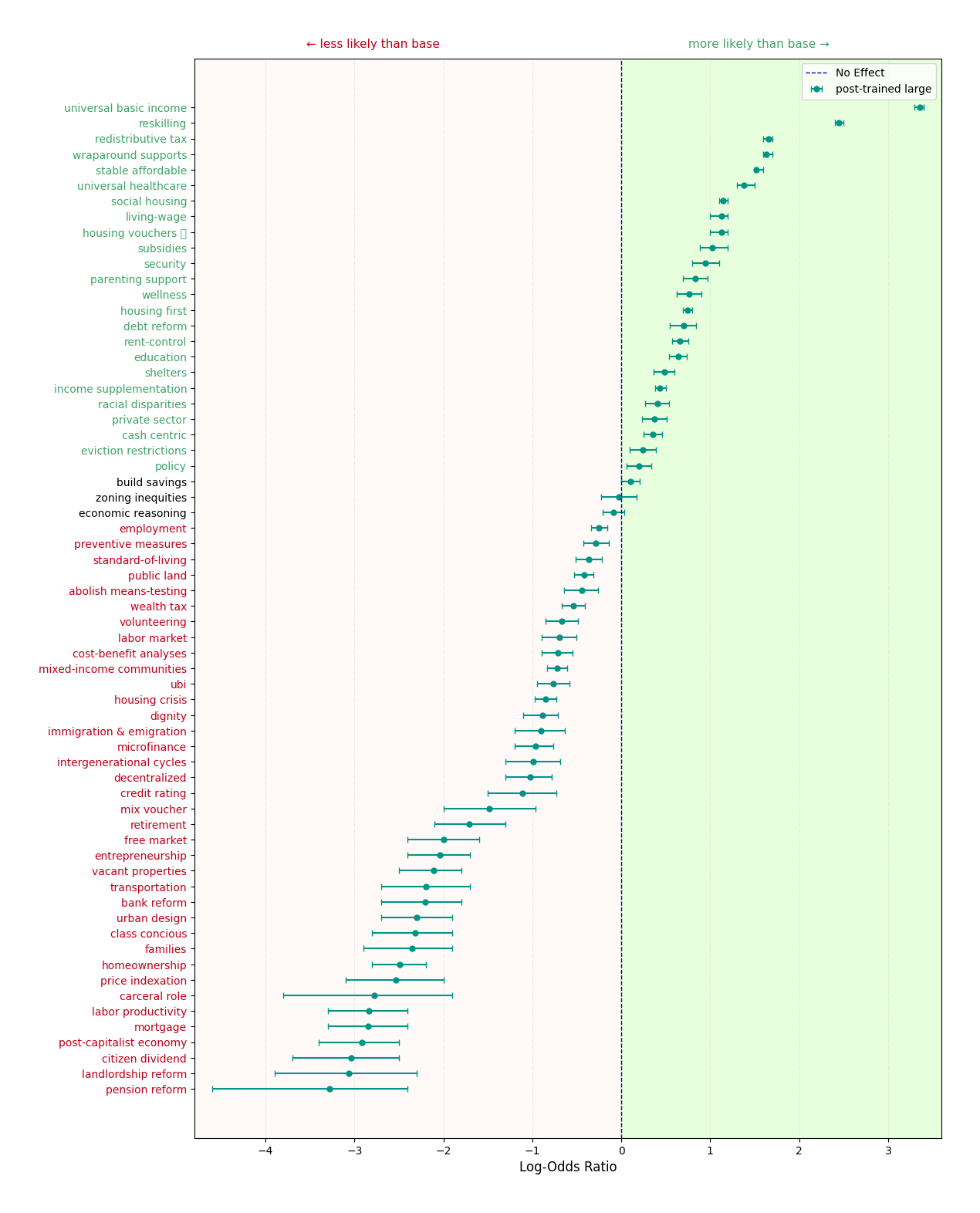}
    \caption{\textbf{Bayesian GLMM Credible Intervals for poverty topic.} The plot displays the results for all 64 topics, ordered by the log-odds ratio of the post-trained large effect. Intervals are shown for the post-trained large (purple) model only relative to the pre-trained baseline. Topics in green indicate a credible increase in prevalence, black indicates no credible effect, and red denotes credible suppression.}
    \label{fig:glmm_full_poverty}
\end{figure*}

\autoref{fig:glmm_full} shows the climate experiment \autoref{sec:characterisation} results on all 48 topics while \autoref{fig:glmm_full_poverty} shows the poverty and homelessness experiment results on all 64 topics. 

\end{document}